\documentclass{article} 
\usepackage{iclr2016_conference,times}
\usepackage{hyperref}
\usepackage{mathtools}
\usepackage{graphicx}
\usepackage{url}
\usepackage[font=small]{caption}
\usepackage{subcaption}
\usepackage[export]{adjustbox}
\usepackage{floatrow}
\newfloatcommand{capbtabbox}{table}[][\FBwidth]
\usepackage{blindtext}
\usepackage{tikz}
\usetikzlibrary{tikzmark}
\usepackage{enumitem}
\usepackage[compact]{titlesec}

\title{Semi-supervised Tuning from Temporal \\Coherence}

\author{Davide Maltoni \& Vincenzo Lomonaco
\\
Department of Computer Science and Engineering DISI\\
Alma Mater Studiorun - University of Bologna\\
Mura Anteo Zamboni, 7, 40126 Bologna, Italy \\
\texttt{\{davide.maltoni,vincenzo.lomonaco\}@unibo.it} \\
}

\newcommand{\argmax}{\operatornamewithlimits{argmax}}

\begin{document}

\maketitle

\begin{abstract}
Recent works demonstrated the usefulness of temporal coherence to regularize supervised training or to learn invariant features with deep architectures. In particular, enforcing smooth output changes while presenting temporally-closed frames from video sequences, proved to be an effective strategy. In this paper we prove the efficacy of temporal coherence for semi-supervised incremental tuning. We show that a deep architecture, just mildly trained in a supervised manner, can progressively improve its classification accuracy if exposed to video sequences of unlabeled data. The extent to which, in some cases, a semi-supervised tuning allows to improve classification accuracy (approaching the supervised one) is somewhat surprising. A number of control experiments pointed out the fundamental role of temporal coherence.
\end{abstract}

\section{Introduction}

Unsupervised learning is very attractive due to the possibility of easily collecting huge amount of patterns to feed data-hungry deep architectures. It also plays a fundamental role in incremental learning scenarios, where labeled data are often available during the initial training, but not at working time.\\
The use of time coherence, that is enforcing a smooth output change while presenting temporally-closed pattern representations, is the bases of unsupervised learning approaches such as Slow Feature Analysis (SFA) (\cite{wiskott2002slow}) and Hierarchical Temporal Memory (HTM) (\cite{george2009towards}, \cite{schmidhuber1992learning}).\\
Recent works proved the usefulness of temporal coherence to regularize supervised training of deep architectures (\cite{mobahi2009deep}, \cite{weston2012deep}) or to deep learn invariant features (\cite{zou2011unsupervised}, \cite{zou2012deep}, \cite{goroshin2015unsupervised}). In particular \cite{mobahi2009deep} trained a Convolutional Neural Network (CNN) with stochastic gradient descent by using a loss function extended with a temporal coherence term. Their training algorithm iteratively performs three interleaved steps aimed at: $i)$ minimizing the negative log-likelihood; $ii)$ minimizing the network output difference for temporal consecutive video frames; $iii)$ maximizing the network output difference for temporal non-closed video frames. The proposed regularization significantly improves the classification accuracy on \textsf{COIL-100} (\cite{nene1996columbia}) and \textsf{ORL} (\cite{samaria1994parameterisation}).\\
Inspired by \cite{mobahi2009deep} we wondered about the effect of completely removing the supervised component from the loss function, and with some surprise, we experimentally found that an HTM, just mildly trained in supervised manner, can significantly improve its classification accuracy, if successively exposed to video sequences of unlabeled data. This is not obvious since in many application domains, it has been shown that semi-supervised and self-training approaches can lead to dangerous drifts (i.e., when mistakes reinforce themselves). \\
Our scenario of interest is different from the typical semi-supervised learning scenario (\cite{zhu2005semi}) where a small set of labelled data and a larger set of unlabeled data (from the same classes) are available since the beginning. In fact, we assume that the unlabeled data are not available initially and become available (in small batches) only at successive stages, once the system has already been trained and the new data (alone) are used for incremental tuning. This better matches human-like learning scenario involving an initial small amount of direct instruction (e.g. parental labeling of objects during childhood) combined with large amounts of subsequence unsupervised experience (e.g. self-interaction with objects). However it is well known that incremental learning poses extra challenges such as the catastrophic forgetting (\cite{mccloskey1989catastrophic}, \cite{french1999catastrophic}), a manifestation of the stability-plasticity dilemma (\cite{mermillod2013stability}).\\
Our incremental tuning approach exploits temporal coherence as a surrogate supervisory signal. It can be used in conjunction with any architecture whose supervised training loss function includes the desired output vector (e.g., the squared difference loss function). In the simplest version we take the network output at time $t$ as desired output at $t+1$ disregarding of the pattern label. A slightly more sophisticated version (still semi-supervised) performs consistently well in our experiments and, under some conditions, its accuracy approaches the supervised one. \\
In order to verify architecture independence of our approach, a number of tests have been carried out with two quite different deep architectures: HTM and CNN. Since both approaches performed equally well on the initial supervised training experiments, we expected similar performance when dealing with incremental learning and semi-supervised tuning. However, this was not the case in our tests, where our CNN implementation suffered more than HTM the forgetting effect and the lack of supervision. \\
To setup our experiments we needed videos where the objects to recognize smoothly move in front of the camera. In particular, to study incremental learning we needed more video sequences of the same object. Instead of collecting a new dataset we decided to generate video sequences from the well-known and largely used \textsf{NORB} dataset (\cite{lecun2004learning}). Further experimental validations are provided on \textsf{COIL-100} dataset. 

\section{Related work}

Semi-supervised learning (\cite{zhu2005semi}, \cite{chapelle2006semi}) exploits both labeled and unlabeled data to build robust models. In particular, in self-training (\cite{rosenberg2005semi}), a classifier is first trained with a small amount of labeled data and then used to classify the unlabeled data. Typically the most confident unlabeled points, together with their predicted labels, are added to the training set. The classifier is re-trained and the procedure repeated. Our approach can be framed in the semi-supervised learning family since we use labeled data for initial training and unlabeled data (form the same classes) for subsequent tuning. However, as pointed out in the introduction, our approach is incremental and the labeled/unlabeled data are used at different stages to mimic human learning. Therefore, particular care must be taken to control the catastrophic forgetting. A recent work by \cite{goodfellow2013empirical} investigates the extent to which the catastrophic forgetting problem occurs for modern neural networks.\\
In specific application domains semi-supervised learning approaches have been proposed to self-update initial models (or templates): see for example \cite{rattani2009template} for biometric recognition and \cite{matthews2004template} for tracking. Several researchers pointed out, that although the use of unlabeled data can substantially increase the system accuracy and robustness, the risk of drifts is always present. For example, in the context of face recognition, \cite{marcialis2008biometric} reported that even with operations of update procedures at high confidence, the introduction of impostors cannot be avoided. Analogously to many domain specific solutions our approach is incremental and can exploit classification confidence. Temporal coherence has already been exploited for face recognition from video (\cite{franco2010incremental}), but the proposed update solution is domain specific and not easily generalizable as that here introduced. \\
The most related research of this study are the works by \cite{mobahi2009deep} and \cite{weston2012deep} where temporal coherence has been embedded in the semi-supervised training of deep architectures. 
However, in those works unlabeled data are used together with labeled one to regularize the supervised training while we first train a system with labeled data and later we tune it with unlabeled data.\\
The biological plausibility of the computational learning approach here proposed is discussed in \cite{li2008unsupervised} whose authors introduce the term UTL (\emph{Unsupervised Temporal slowness Learning}) to describe the hypothesis under which invariance is learned from temporal contiguity of object features during natural visual experience without external supervision.

\section{SST: Semi-Supervised Tuning}
\label{sec:sst}
Let $S_w$ be a temporally coherent sequence of video frames $v^{(t)},t=1 \dots len(S_w)$ taken from the same object (of class $w$): while the total object variation (in term of pose, lighting, distortion, etc.) in the whole sequence can be very high, only a limited amount of variation is expected to characterize pairs of successive frames $v^{(t)}$ and $v^{(t-1)}, t=2 \dots len(S_w)$.\\
Let $N$ be a classifier able to map an input pattern $v^{(t)}$  (i.e., a single video frame) into an output vector $N(v^{(t)})$ denoting the posterior class probabilities $P(w | v^{(t)}), w=1 \dots n_w$. While in this work $N$ will be instantiated with a deep architecture trained with gradient descent, in general $N$ can be any trainable classifier returning class probabilities and whose optimization procedure minimize a cost (or loss) including the desired output $d(v^{(t)})$ for the input $v^{(t)}$. \\
If the squared error is taken as loss function, for each pattern $v^{(t)}$ (of class $w$) the optimization procedure attempts to minimize:

\begin{equation}
\label{eq:minerror}
\frac{1}{2} \left \| N(v^{(t)})-d(v^{(t)}) \right \|^2
\end{equation}

Assuming that $N$ has already been trained (with supervision) by using a first batch of data, each subsequence training can be considered as a tuning (i.e., we start with learned parameters). Given a sequence $S_w$, 
we define four ways to instantiate the desired vector $d(v^{(t)})$ during the system tuning: 

\begin{itemize}[leftmargin=.7cm]
\item \emph{Supervised Tuning} (\textit{\textbf{SupT}}): this is the classical supervised approach where the desired output vector has the $\Delta$ form (all terms are zero except that corresponding to the pattern class $w$) 
\begin{equation}
d(v^{(t)})=\Delta_w=[0,\dots,1,\dots,0]
\begin{tikzpicture}[remember picture,overlay]
\draw[<-] 
  ([shift={(-33pt,-5pt)}]pic cs:a) |- ([shift={(-50pt,-10pt)}]pic cs:a) 
  node[anchor=east] {$\scriptstyle w$};  
\end{tikzpicture}
\end{equation}
\vspace{.1cm}
\item \emph{Supervised Tuning with Regularization} (\textit{\textbf{SupTR}}): 
\begin{equation}
d(v^{(t)})=\lambda \cdot \Delta_w + (1-\lambda) \cdot N(v^{(t-1)})
\end{equation}
where $\lambda \in [0,1]$ controls the influence of the temporal coherence regularizing term. This is close to the approach proposed by \cite{mobahi2009deep}, but we embed the regularizing term into the desired output and then perform a single optimization step, while \cite{mobahi2009deep} make disjoint optimization steps.
\item \emph{Semi-Supervised Tuning - Basic} (\textit{\textbf{SST-B}}): 
\begin{equation}
d(v^{(t)}) = N(v^{(t-1)})
\end{equation}
This simply takes as desired output at time $t$ the output vector at time $t-1$. The class label $w$ is not used, but since we assume that the input pattern belongs to one of the know-classes, the update is semi-supervised. 
\item \emph{Semi-Supervised Tuning - Advanced} (\textit{\textbf{SST-A}}):
\begin{equation}
f(v^{(t)}) = \begin{dcases}
N(v^{(t-1)}) & t=2\\ 
\frac {f(v^{(t-1)})+N(v^{(t-1)})}{2} & t>2
\end{dcases}
\end{equation}
\begin{equation}
d(v^{(t)}) = \begin{dcases*}
N(f^{(t)}) & if $\max \limits_{i} f_i(v^{(t)}) > sc$\\ 
N(v^{(t)}) & otherwise
\end{dcases*}
\end{equation}
At each step, we fuse the posterior probabilities $N(v^{t-1})$ with the posterior probabilities $f(v^{(t-1)})$ accumulated before; this is a sort of sum rule fusion where the weight of far (in time) patterns progressively vanishes. Then, if at least one of the fused class posteriors (in $f(v^{(t)})$) is higher than a given threshold $sc$, denoting high self-confidence, the desired output is set to $f(v^{(t)})$ to enforce temporal coherence. Otherwise (high uncertainty cases) no semi-supervised update have to be done, and formally, this can be achieved by passing back $N(v^{(t)})$ to equation (\ref{eq:minerror}). Here too, the class label $w$ is not used.
\end{itemize}

\section{Dataset and Architectures}

\subsection{Norb revisited}
\label{sec:norb_revisited}
Instead of collecting just another dataset we focused on the well know and largely used \textsf{NORB} dataset (\cite{lecun2004learning}). This is still one of the best dataset to study invariant object recognition and well-fit our purposes because it contains 50 objects and 972 variations for each objects. The 50 objects belong to 5 classes (10 objects per class) and the 972 variations are produced by systematically varying the camera elevation (9 steps), the object azimuth with respect to the camera (18 steps) and the lighting condition (6 steps). 

\begin{figure}[h]
\RawFloats
\floatbox[{\capbeside\thisfloatsetup{capbesideposition={left,top},capbesidewidth=.1cm}}]{figure}[\FBwidth]
{\caption*{a)}}
{\includegraphics[width=0.95\textwidth]{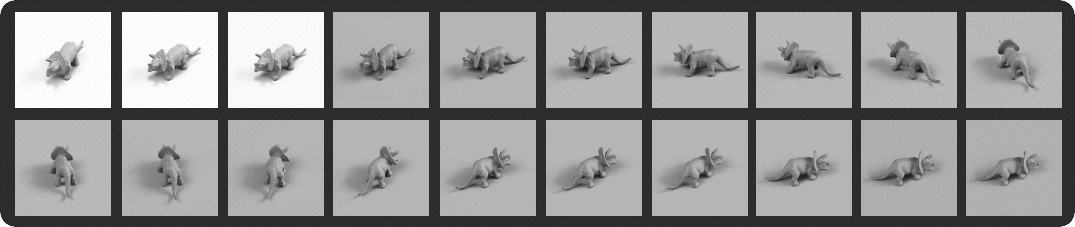}}
\vspace{.2cm}
\floatbox[{\capbeside\thisfloatsetup{capbesideposition={left,top},capbesidewidth=.1cm}}]{figure}[\FBwidth]
{\caption*{b)}}
{\includegraphics[width=0.95\textwidth]{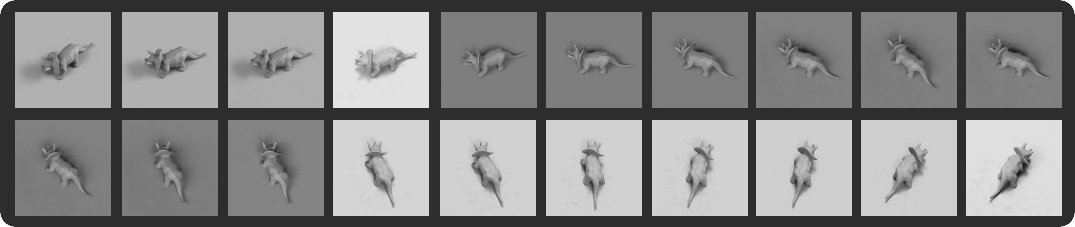}}
\caption{ \small a) An example of training sequence (20 frames). b) A test sequence with \emph{mindist} = 4 from the previous training sequence.}
\label{fig:seq_examples}
\end{figure}

Temporally coherent video sequences can be generated from \textsf{NORB} by randomly walking the 3d (elevations, azimuth, lighting) variation space, where consecutive frames are characterized by a single step along one dimension. In our generation approach the random walking is controlled by some parameters like the number of frames, the probability of taking a step along each of the 3 dimensions, the probability of inverting the direction of movement (flip back), etc. Fig. \ref{fig:seq_examples}.a shows an example of training sequence. When generating test sequences we must avoid to include frames already used in the training sequences. In particular, when generating test sequences (with a given \emph{mindist}), we ensure that each test frame has a city-block distance of at least mindist steps (\emph{mindist} $\geq$ 1) from any of the training set frames. Fig. \ref{fig:seq_examples}.b shows a test sequence with \emph{mindist} = 4 with respect to the training sequence of Fig. \ref{fig:seq_examples}.a. 

In the standard \textsf{NORB} benchmark for each of the 5 classes, 5 objects are included in the training set and 5 objects in the test set. In the proposed benchmark we prefer to focus on pose and lighting invariance hence our training and test set are not differentiated by the object identity but by the object pose and lighting (for an amount modulated by \emph{mindist}). However, for completeness, in Appendix \ref{app:norb_native} we also report results on an equivalent benchmark where the native object segregation is maintained. In our benchmark we also focus on monocular representation since the availability of stereo information makes the problem unnecessarily simpler for the task at hand. The benchmark dataset used in our experimentation consists of:
\begin{itemize}[leftmargin=.7cm]
\item 10 training batches $TrainB_i$. Each $TrainB_i$ is 1,000 frames wide and is composed by 50 temporally coherent sequences (20 frames wide), each representing one of the 50 objects. $TrainB_1$ is used for initial training and $TrainB_2, \dots, TrainB_{10}$ for successive incremental tuning. When training the system on $TrainB_i$ we have no longer access to the previous $TrainB_j, j<i$. We do not enforce any \emph{mindist} among training set sequences, so the same frame can be present in different batches.
\item 10 test batches $TestB_i$ for each \emph{mindist} = 1, 2, 3 and 4. Test batches are structured as the train batches, but here \emph{mindist} is enforced, so each frame included in the test batches has a distance of at least \emph{mindist} from the 10$\times$1,000 frames\footnote{Actually due to the presence of duplicates in our training random walks, the number of different frames is 8,531 (smaller  than 10,000).} of the training batches. Higher \emph{mindist} values make the classification problem more difficult, because patterns are less than similar with respect to the training set ones. The temporal coherent organization of the test batches allows two type of evaluations to be performed:
\begin{itemize}[leftmargin=.7cm]
\item \emph{Frame based classification}: here temporal organization is not considered and each frame has to be classified independently of its sequence/positions in the batch. For simplicity, for each \emph{mindist} we can treat the batches $TestB_i, i=1,\dots,10$, as as a single plain test set of 10$\times$1,000 patterns.
\item \emph{Sequence based classification}: this evaluation (not included in the experiments carried out in this paper) is aimed at classifying sequences and not single frames, so one can exploit multiple frames per object and their temporal coherence. Of course this is a simpler classification problem due to the possibility of fusing information. As side effect the number of pattern to classify reduces to $\frac{10,000}{20} = 500$. 
\end{itemize}
\end{itemize} 
With the purpose of evaluating our approach on a harder problem we can consider another benchmark (denoted as the “50-class benchmark”) where each object is considered as a separate class. It is worth noting that this is a quite complex problem due to the sometime small variability among objects originally belonging to the same class. To setup this benchmark we can still use the above sequences, with the only caution of ignoring original class labels and taking object labels as class labels.\\
Original \textsf{NORB} images are 96$\times$96 pixels. We noted that working on reduced resolution images (up to 32$\times$32) does not reduce classification accuracy (on the 5-class problem). So in order to speed-up the experiments we downsampled the \textsf{NORB} images to 32$\times$32 pixels\footnote{The same downsampling was done in other works (\cite{saxe2011random}, \cite{ngiam2010tiled}, \cite{wagner2013learning}).}\\
The full training and test sequences used in this study (provided as sequences of filenames referring to the original \textsf{NORB} images) can be downloaded from [\url{https://bitbucket.org/vincenzo_lomonaco/norbcreator}]. In the same repository we make available the tool (and the code) used to generated the sequences.

\subsection{HTM}
\begin{figure}[h]
\begin{center}
\includegraphics[width=.65\linewidth]{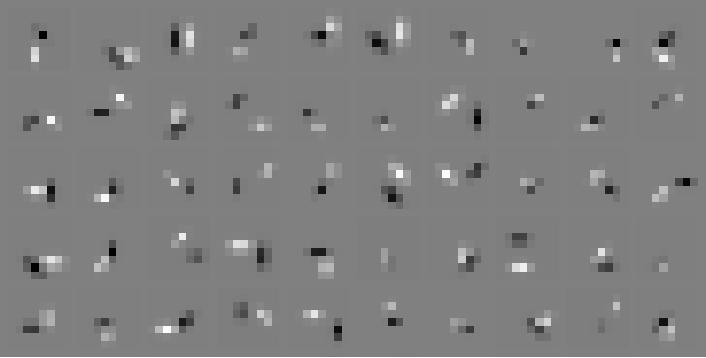}
\caption{\small A graphical representation of the 50 ``random'' dilobe filters at level 1.}
\label{fig:dialobe}
\end{center}
\end{figure}

Hierarchical temporal memory (HTM) (\cite{george2009towards}) is a biologically inspired framework that can be framed into multistage Hubel-Wiesel architectures (\cite{ranzato2007unsupervised}), a specific family of deep architectures. A brief overview of HTM is provided in Appendix \ref{app:htm_overview}. A more comprehensive introduction can be found in \cite{maltoni2011pattern} and \cite{rehn2014incremental}.\\
Analogously to CNN, HTM hierarchical structure is composed of alternating feature extraction and feature pooling layers. However, in HTM feature pooling is more complex than typical sum o max pooling used in CNN, and the time is used since the first training steps, when HTM self-develops its internal memories, to form groups of feature detectors responding to temporally-close inputs. This is conceptually similar to the unsupervised feature learning proposed in \cite{zou2011unsupervised}, \cite{zou2012deep}, \cite{goroshin2015unsupervised}. \\ 
In the classical HTM approach (\cite{maltoni2011pattern}) once a network is trained, its structure is frozen, thus making further training (i.e., incremental learning) quite critical. \cite{rehn2014incremental} introduced a technique (called HSR) for HTM (incremental) supervised training based on gradient descent error minimization, where error backpropagation is implemented through native HTM message passing based on belief propagation. In the present work HSR will be used for semi-supervised tuning.\\
The HTM architecture here adopted includes some modifications with respect to \cite{maltoni2011pattern}, \cite{rehn2014incremental}. We experimentally verified that these modifications yield to some accuracy improvement when working with natural images, and, at the same time, reduce the network complexity. Presenting these architectural changes in detail is not in the scope of this paper, but some hints are given in the following:
\begin{itemize}[leftmargin=.7cm]
\item \emph{Dilobe ordinal filters}: the feature extraction at the first level is not based on a variable number of self-learned templates as described in \cite{maltoni2011pattern}, but is carried out with a bank of 50 dilobe ordinal filters \cite{sun2009ordinal}. Each filter, of size 8$\times$8, is the sum of two 2d Gaussians (one positive and one negative) whose center, size and orientation are randomly generated (see Fig.\ref{fig:dialobe}). Each filter computes a simple intensity relationship between two adjacent regions (i.e. the two filter lobes) which is quite robust with respect to light changes, and (in our experience) discriminant enough for low level feature extraction. Although one could setup an unsupervised approach to learn optimal filters from natural images, for simplicity in this work we generated them randomly and kept them fixed.
\item \emph{Partial receptive field}. in the classical HTM implementation the receptive field of a node is the union of the child node receptive fields, and is not possible for a node to focus only on a specific portion of its receptive field. In general, this does not allows to isolate objects from the background or to deal with partial occlusions. A simple but effective technique has been implemented to deal with partial receptive fields.
\end{itemize}

The HTM architecture used in our experiments has 5 levels:
\begin{itemize}[leftmargin=.7cm, noitemsep]
\item Input: 1024 nodes (32$\times$32) connected to image pixels.
\item Intermediate 1: 169 (13$\times$13 nodes), each node has 8$\times$8 child nodes. 
\item Intermediate 2: 169 (13$\times$13 nodes), each node has 1 child node.
\item Intermediate 3: 9 (3$\times$3 nodes), each node has 5$\times$5 child nodes.
\item Output: 1 node with 3$\times$3 child nodes.
\end{itemize}

Since intermediate level 2 and 3 include both feature extraction and feature pooling, and the input level in this case is not performing any particular processing, the above 5 levels correspond to 6 levels in a CNN (accidentally the same of LeNet7). Note that an HTM node is much more complex than a single artificial neuron in conventional NN, since it was conceived to emulate a cortical micro-circuit (\cite{george2009towards}). 
HTM accuracy on baseline \textsf{NORB} is reported as additional material in Appendix \ref{app:baseline_acc}.

\subsection{CNN: LeNet7 on Theano}
\label{sec:lenet7_on_theano}
\begin{figure}[h]
\begin{center}
\includegraphics[width=.8\linewidth]{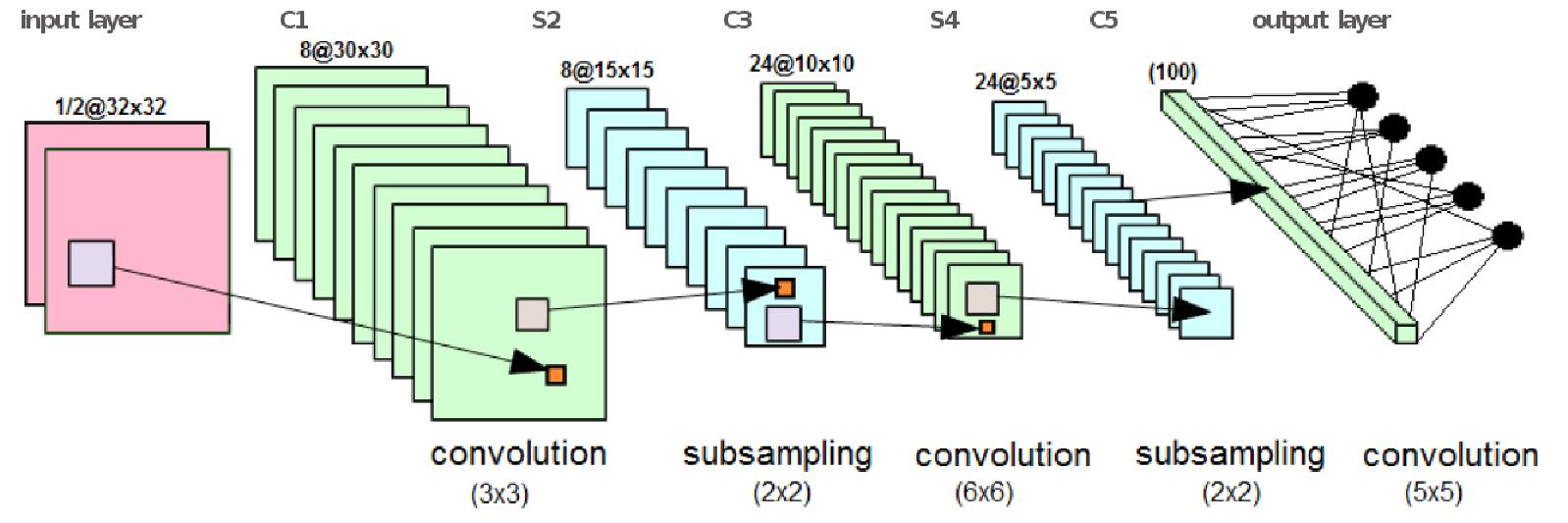}
\caption{\small The CNN used in this work (original LeNet7 adapted to 32$\times$32 inputs). X@Y$\times$Y stands for X feature maps of size Y$\times$Y; (Z$\times$Z) stands for the filters of size Z$\times$Z.}
\label{fig:lenet7}
\end{center}
\end{figure}

The CNN architecture used in our experiments is an minor modification of ``LeNet7'' that was specifically designed by \cite{lecun2004learning} to classify \textsf{NORB} images. This is still one of the best performing architecture on \textsf{NORB} benchmarks. We empirically proved that working on 32$\times$32 images does not reduce accuracy with respect to the 96$\times$96 original images. So our main modification concerns the reduced feature map size and filter size to deal with 32$\times$32 monocular inputs (see Fig. \ref{fig:lenet7}).

\cite{lecun2004learning} suggested to train LeNet7 with the squared error loss function, which naturally fits our semi-supervised tuning formulation. In our experiment on the standard \textsf{NORB} benchmark we evaluated some modifications to the architecture or the training procedure: $i)$ Max pooling instead of the original sum pooling; $ii)$ Soft-max + log-likelihood instead of squared error; $iii)$ Dropout; but none of these changes (nowadays commonly used to train CNN on large datasets) led to consistently better accuracy, so we came back to the original version that was easily implemented in Theano (\citep{bergstra2010theano}).
Since we are not using any output normalization\footnote{Any attempt to introduce a normalization (e.g. soft-max) resulted in some accuracy loss.}, the network output vector is not exactly a probability vector: however looking at the output values we noted that after a few training iterations they approximate quite well a probability vector: all elements in $[0,1]$ and summing to $1$.  
To be sure of the soundness of our CNN implementation and training procedure we tried to reproduce the results reported in \cite{lecun2004learning} for a LeNet7 trained on the full ``normalized-uniform'' dataset of 24300 patterns (4860 for each of the 5 classes). Since results in \cite{lecun2004learning} are reported only for the binocular case, for this control experiment we also used binocular patterns (even if in the format 32$\times$32). After some tuning of the training procedure, we achieved a classification error of 5.6\% which is slightly better than the 6.6\% reported in \cite{lecun2004learning}, aligned with 5.6\% of \cite{ranzato2007unsupervised} and not far from the state-of-the-art 3.9\% reported in \cite{ngiam2010tiled}. More details on the CNN accuracy on baseline \textsf{NORB} (including a comparison with HTM) can be found in Appendix \ref{app:baseline_acc}.

\section{Experiments}
\label{sec:experiments}
\subsection{Incremental tuning}
\label{sec:inc_tun}
In this section we focus on incremental learning and evaluate the (4) tuning strategies introduced in section \ref{sec:sst} on the benchmark proposed in section \ref{sec:norb_revisited}. In all the experiments:
\begin{itemize}[leftmargin=.7cm]
\item We used 32$\times$32 monocular patterns (left eye only).
\item We report classification accuracy as \emph{frame based classification accuracy} (the \emph{sequence based classification} scenario will be addressed in future studies). See section \ref{sec:norb_revisited} for the definition of the two scenarios.

\item For semi-supervised tuning, each training batch of 1,000 frames is treated as a single frame flow, without exploiting the regular sequence order and size within the batch to isolate the 50 temporally-coherent sequences. In fact, even if in natural vision abrupt gaze shifts could be detected to segment sequences, we prefer to avoid simplifying assumptions on this.

\item To limit bias induced by the batch order presentation, we averaged experiments over 10 runs and at each run we randomly shuffled the batches $TrainB_i, i=2,\dots,10$ ($TrainB_1$ is always used for initial supervised training). By measuring the standard deviation across the 10 runs, we can also study the learning process stability.
\item To avoid overfitting we did not performed a fine adjustments of parameters characterizing the (parametric) update strategies. We set them according to some exploratory tests and then kept the same values for all the experiments:
\begin{itemize}[leftmargin=.7cm]
\item For \textit{SupTR}, the weight $\lambda$ of the supervised component is set to $\frac{2}{3}$.
\item For \textit{SST-A} the self-confidence threshold $sc$ is set to 0.65.
\end{itemize}
\end{itemize}

When performing incremental learning, care must be taken to avoid catastrophic forgetting. In fact, since patterns belonging to previous batches are no longer available, training the system with new patterns could lead to forget old ones. Even if in our tuning scenario the new patterns come from the same objects (pose and lighting variations) and there is some overlapping\footnote{Since we are not enforcing any \emph{mindist} between training sequences, the random walk can lead to the inclusion of the same frame in different sequences/batches. In our opinion, this better emulates unsupervised human experience with objects, where the same object view can be refreshed over time.} in the training sequences, catastrophic forgetting is still an issue. \\
For HTM we experimentally found that a good tradeoff between stability and plasticity can be achieved by running only 4 HSR iterations for each batch of 1,000 patterns, while for CNN we found that the optimal number of iterations is much higher (about 100 iterations).\\~\\


\begin{figure}[h]
\center
\begin{subfigure}{.5\textwidth}
	\includegraphics[width=0.95\linewidth]{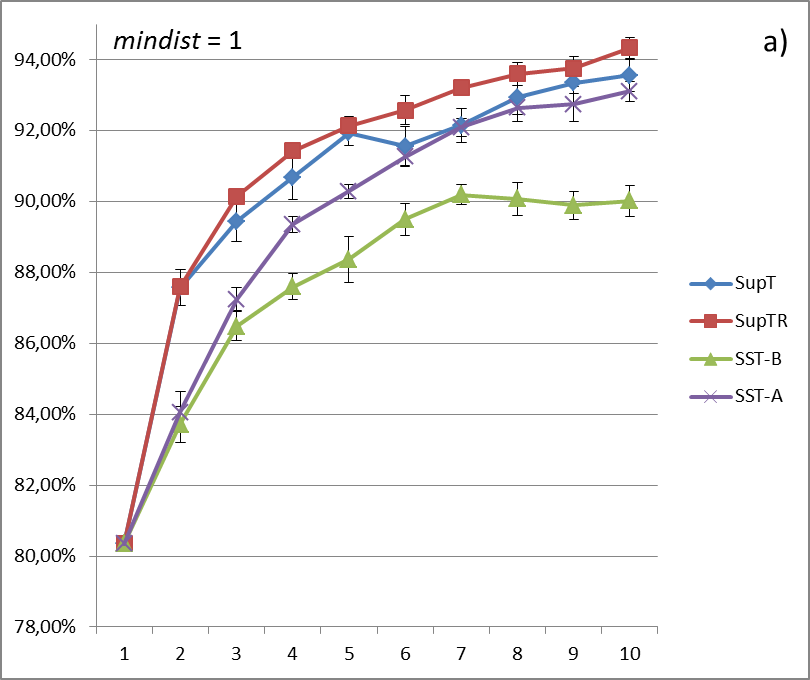}

\end{subfigure}%
\begin{subfigure}{.5\textwidth}
	\includegraphics[width=0.95\linewidth]{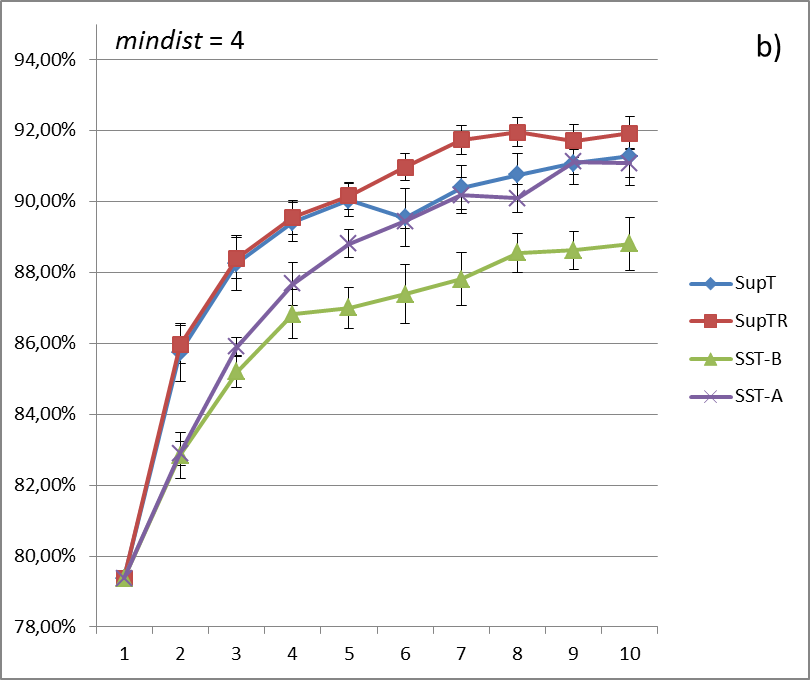}

\end{subfigure}
\caption{\small HTM accuracy on the test set for \emph{mindist} = 1 (a) and \emph{mindist} = 4 (b). Positions $2, \dots, 10$ (x-coordinate) denote the test set accuracy after incremental tuning with batches $TrainB_i, i=2,\dots,10$. Position $1$ exhibits the same accuracy for all the update strategies because it denotes the accuracy after supervised training on $TrainB_1$. The bars denote 95\% mean confidence intervals (over 10 runs).}
\label{fig:htm_acc}
\end{figure}

Fig. \ref{fig:htm_acc} shows HTM accuracy at the end of pre-training\footnote{No additional data (e.g., jittered patterns) is used for HTM pre-training.} on $TrainB_1$ (point 1) and after each incremental tuning (points $2,\dots,10$). We note that:
\begin{itemize}[leftmargin=.7cm]
\item Supervised tuning \textit{SupT} works well and each new batch of data contributes to increase the overall accuracy.
\item Regularized supervised tuning \textit{SupTR} performs slightly better than \textit{SupT}, and more important, makes the learning process more stable; this can be appreciated by the smoother trend in the graphs and by the average standard deviation over the 10 runs, that (for \emph{mindist} = 1) is 0.7\% for SupT and 0.4\% for \textit{SupTR}. This is in line with results of \cite{mobahi2009deep}, where a relevant accuracy improvement was reported on \textsf{COIL-100} when regularizing the supervised learning with temporal coherence. Here the gap between \textit{SupT} and \textit{SupTR} is smaller than in \cite{mobahi2009deep}, probably due to the fact that our tuning batches are quite small (1,000 patterns) and regularization plays a minor role.
\item \textit{SST-B} and \textit{SST-A} accuracy is surprisingly good when compared with supervised accuracy, proving that temporal continuity is a very effective surrogate of supervision for HTM. Initial trends of \textit{SST-B} and \textit{SST-A} are similar, then \textit{SST-B} tends to stabilize while \textit{SST-A} accuracy continues to increase approaching supervised update \textit{SupT}. The self-confidence computation that \textit{SST-A} uses to decide whether updating the gradient or not, seems to be a valid instrument to skip cases where temporal continuity is not effective (e.g. change of sequence, very ambiguous patterns, etc.).
\end{itemize}

\begin{figure}[h]
\center
\begin{subfigure}{.5\textwidth}
	\includegraphics[width=0.95\linewidth]{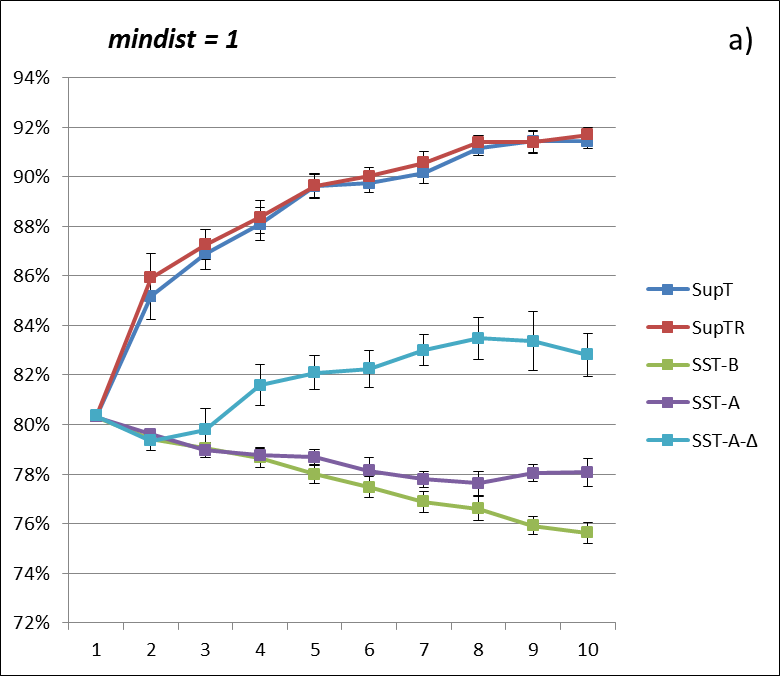}

\end{subfigure}%
\begin{subfigure}{.5\textwidth}
	\includegraphics[width=0.95\linewidth]{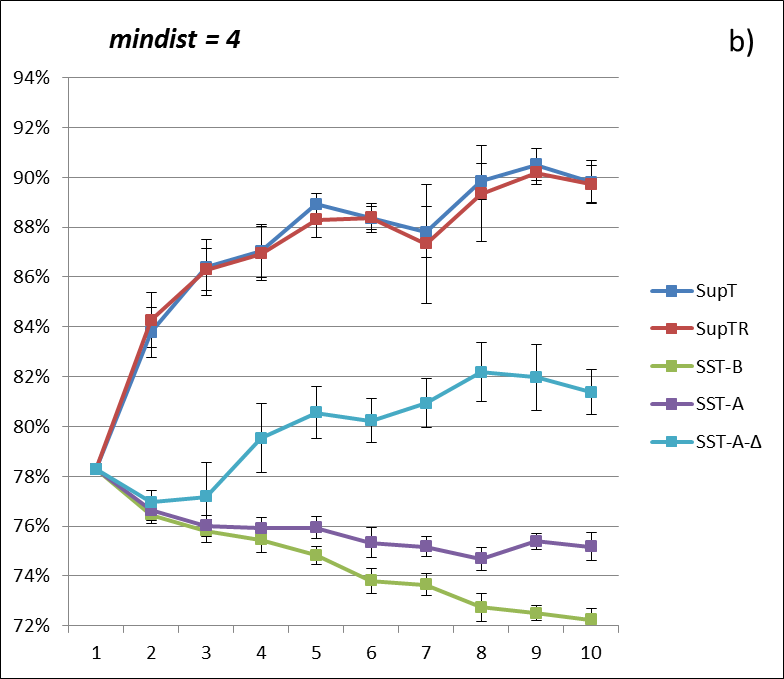}

\end{subfigure}
\caption{\small CNN accuracy on the test set for \emph{mindist} = 1 (a) and \emph{mindist} = 4 (b).}
\label{fig:cnn_acc}
\end{figure}

Fig. \ref{fig:cnn_acc} shows the results of the same experiment performed with CNN. Here we observe that:
\begin{itemize}[leftmargin=.7cm]
\item Accuracy at the end of initial supervised training (on $TrainB_1$) is similar to HTM. 
\item \textit{SupT} and \textit{SupTR} lead to a remarkable accuracy improvement during incremental tuning with $TrainB_i, i=2,\dots,10$ even if accuracy is about 2\% lower than HTM and for \emph{mindist} = 4 the learning process appears to be less stable.
\item Unexpectedly, the semi-supervised tuning \textit{SST-B} and \textit{SST-A} did not work with our CNN implementation. We tried some modifications (architecture, learning procedure) but without success. The only way we found to increase accuracy in the semi-supervised scenario is with the variant of \textit{SST-A} (denoted as \textit{SST-A-$\Delta$}) introduced and discussed in section \ref{sec:control_exp}, However, also for \textit{SST-A-$\Delta$} the accuracy gain is quite limited if compared with semi-supervised tuning on HTM.
\end{itemize}
A similar trend can be observed in the experimental results reported as additional materials (Appendix \ref{app:norb_native}), where the native object segregation is maintained.


\subsection{Making the problem harder}
The good performance of HTM in semi-supervised tuning reported in the previous section could be attributed to the initial high-chance of self-discovering the pattern class. In fact, if the initial classification accuracy is high enough, the missing class label can be replaced by a good guess. To study SST effectiveness for harder problems, where the initial classification accuracy is lower, we set-up two experiments:
\begin{itemize}[leftmargin=.7cm]
\item the former consists in deliberately (and progressively) deteriorating the initial classification accuracy by providing a certain amount of wrong labels during the supervised training on $TrainB_1$.
\item the latter uses the same training and test batches but turns the problem into a 50-class classification. As discussed in section \ref{sec:norb_revisited}, this is much more difficult (expecially for 32$\times$32 patterns) because different \textsf{NORB} objects (e.g. two cars) are visually very difficult to distinguish at certain angles (even for humans).
\end{itemize}

\begin{figure}[h]
\center
\begin{subfigure}{.5\textwidth}
	\includegraphics[width=0.95\linewidth, height=5.05cm]{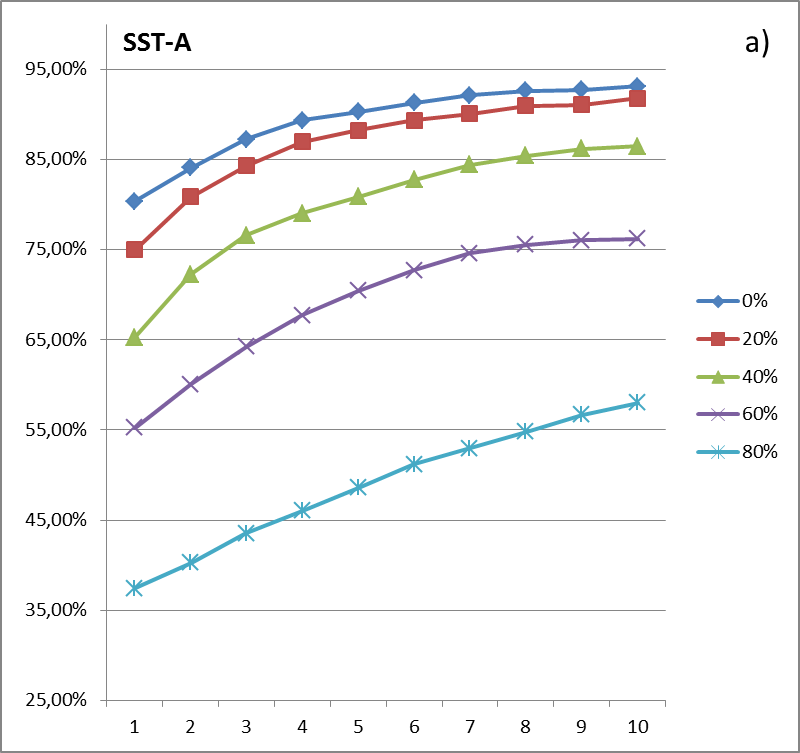}

\end{subfigure}%
\begin{subfigure}{.5\textwidth}
	\includegraphics[width=0.95\linewidth]{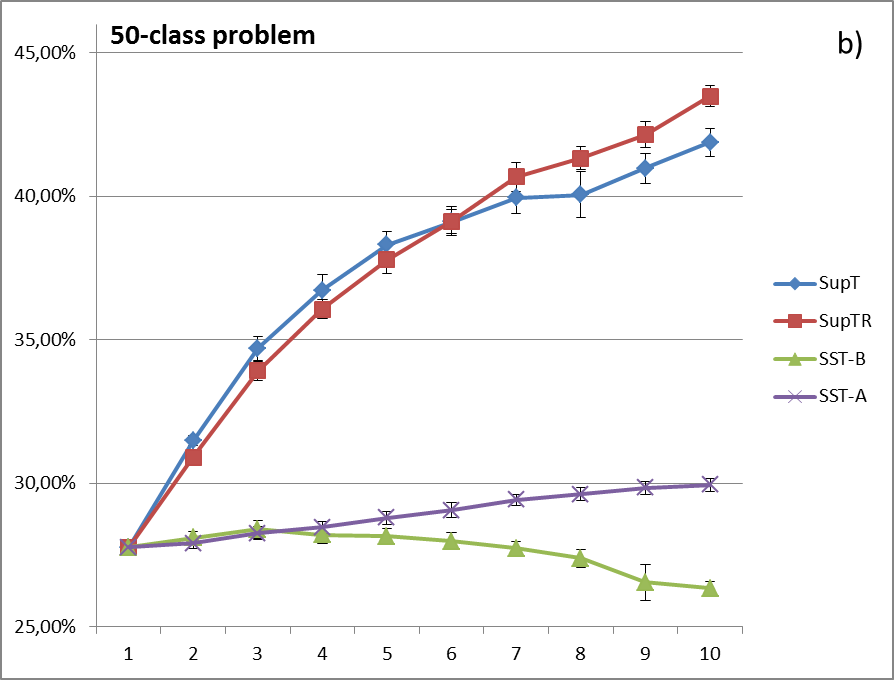}

\end{subfigure}
\caption{a) HTM + \textit{SST-A} accuracy on the test set (\emph{mindist} = 1) for different amounts of wrong labels provided during initial supervised training on $TrainB_1$. b) HTM accuracy on the test set (\emph{mindist} = 1) for different update strategies on the 50-class problem.}
\label{fig:prob_harder}
\end{figure}

Fig. \ref{fig:prob_harder} shows results of these experiments. We note that:
\begin{itemize}[leftmargin=.7cm]
\item As the initial classification accuracy degrades, \textit{SST-A} accuracy degrades gently and the gap between initial and final accuracy remains high. Even a limited initial accuracy of about 35\% does not prevent \textit{SST-A} to benefit from semi-supervised tuning.
\item Of course here the gap between \textit{SST-A} and supervised tuning \textit{SupT}/\textit{SupTR} (not reported in the graph) is higher because supervised tuning is able to overcome the introduced initial degradation since the second batch, always leading to a final accuracy close to Fig. \ref{fig:htm_acc}.a.
\item The 50-class experiment can be considered an extreme case, because the initial classification accuracy is about 25\% and even supervised tuning approaches (\textit{SupT} and \textit{SupTR}) are not able to increase final performance over 44\%. In this scenario \textit{SST-B} after an initial stability (batches $2,\dots,5$) starts drifting away (batches $6,\dots,10$). On the contrary, \textit{SST-A} denotes a stable (even if limited) accuracy gain, proving to be able to operate also in high uncertainty conditions. 
\end{itemize}


\subsection{Control experiments}
\label{sec:control_exp}
In this section we introduce further experiments with the aim to better understand the factors contributing to the success of semi-supervised tuning. In particular we modified \textit{SST-A} as:
\begin{itemize}[leftmargin=.7cm]
\item \textit{\textbf{SST-A-$\Delta$}}:
\begin{equation}
d(v^{(t)}) = \begin{dcases*}
\Delta_{\argmax \limits_i f(v^{(t)})} & if $\max \limits_i f_i(v^{(t)})>sc$\\ 
N(v^{(t)}) & otherwise
\end{dcases*}
\end{equation}
This is very similar to \textit{SST-A}, in fact $f(v^{(t)})$  is computed in the same way by exploiting temporal coherence, but here when the self-confidence is higher than the threshold, instead of enforcing the temporal coherent pattern $f(v^{(t)})$, we pass back the delta distribution corresponding to the self-guessed class. 
\item \textit{\textbf{SST-A-$\Delta$-noTC}}:
\begin{equation}
d(v^{(t)}) = \begin{dcases*}
\Delta_{\argmax \limits_i N_i(v^{(t)})} & if $\max \limits_i N_i(v^{(t)})>sc$\\ 
N(v^{(t)}) & otherwise
\end{dcases*}
\end{equation}
Here no temporal coherence is used neither for estimating self-confidence nor for enforcing output continuity. This correspond to the basic self-training approach used in several applications.
\end{itemize}

\begin{figure}[h]
\center
\begin{subfigure}{.5\textwidth}
	\includegraphics[width=0.95\linewidth]{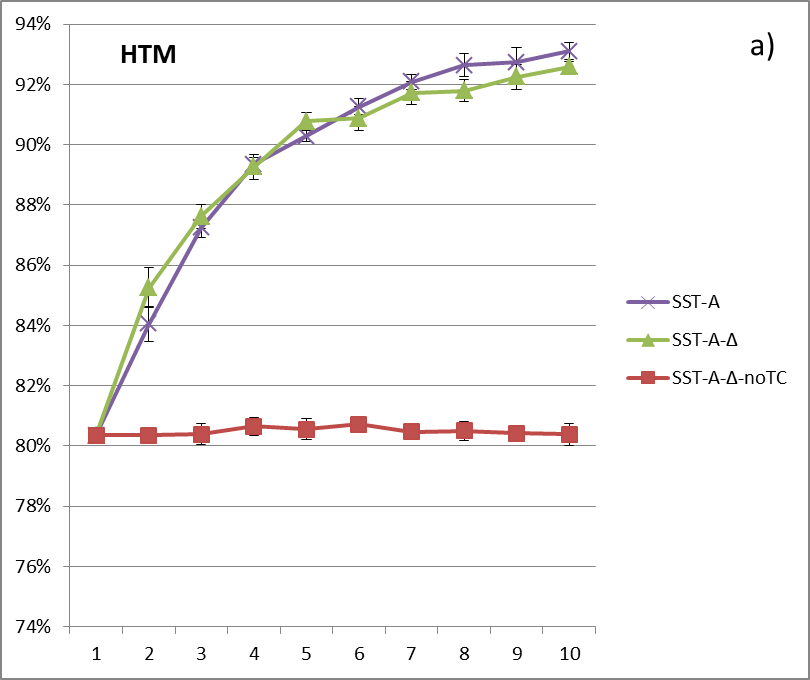}

\end{subfigure}%
\begin{subfigure}{.5\textwidth}
	\includegraphics[width=0.95\linewidth, height=5.56cm]{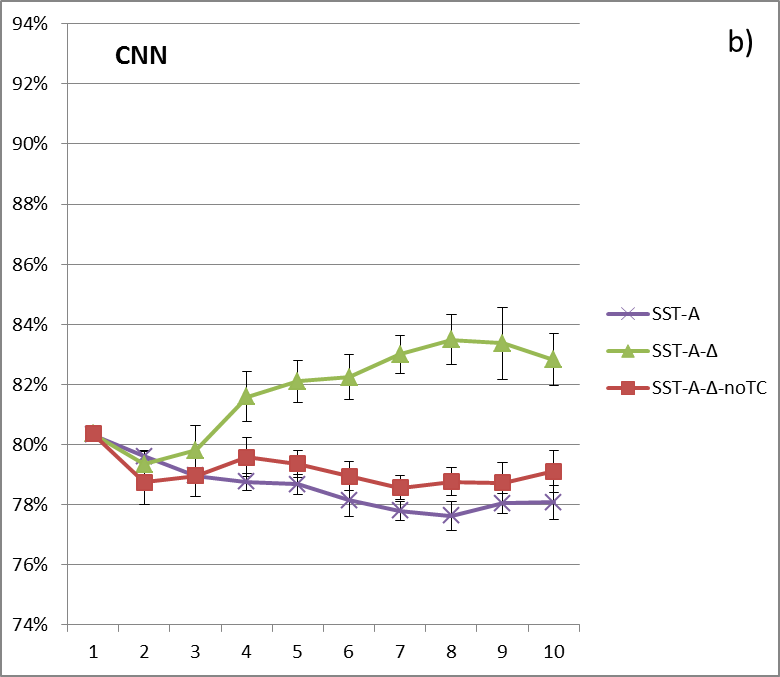}
	
\end{subfigure}
\caption{a) HTM accuracy (5-class problem, \emph{maxdist} = 1) on SST-A and its two variants. b) CNN accuracy (5-class problem, \emph{maxdist} = 1) on SST-A and its two variants. }
\label{fig:control_exp}
\end{figure}

Fig. \ref{fig:control_exp}.a compares HTM accuracy on \textit{SST-A} and the two above variants. The small gap between \textit{SST-A} and \textit{SST-A-$\Delta$} (in favor of SST-A) can be attributed to the regularizing effect of passing back a temporally coherent output vector instead of a sharp delta vector. A totally unsatisfactory behavior can be observed for the second variant (\textit{SST-A-$\Delta$-noTC}) where the network cannot look back in time but can only exploits the current pattern: the flat accuracy in the graph testifies that in this case self-training does not allow HTM to improve. This is a classical pitfall of basic self-training approaches where the patterns whose label can be correctly guessed do not bring much value to the improve the current representation while really useful patterns (in term of diversity) are not added because of the low self-confidence.

Fig. \ref{fig:control_exp}.b shows CNN accuracy for the same experiments. While \textit{SST-A-$\Delta$-noTC} here too remains ineffective, in this case \textit{SST-A-$\Delta$} is much better than \textit{SST-A}, even if far from semi-supervised accuracy achieved by HTM. But why our CNN implementation does not tolerate a desired output vector made of (combinations) of past output vectors, and prefer a more radical delta vector computed by self-estimation of the pattern class? By comparing the output vectors produced by HTM and CNN when making inference on new patterns, we noted that HTM posterior probabilities are quite peaked around one class (similarly to delta form) while for CNN they are more softly spread among different classes. Numerically this can be made explicit by computing the average entropy over the network outputs of 1,000 previously unseen patterns: for CNN we measured and entropy of 1.44 bit, while for HTM the entropy is 0.50 bit, which is much closer to the 0 entropy of delta vector. Therefore, it seems that HTM output vectors are already in the right form for the loss function, while CNN output vectors need to be sharpened to make learning more effective.


\subsection{Further experimental validation on COIL-100}
\label{app:exp_coil100}
\begin{figure}[h]
\center
\begin{subfigure}{.5\textwidth}
	\includegraphics[width=0.95\linewidth]{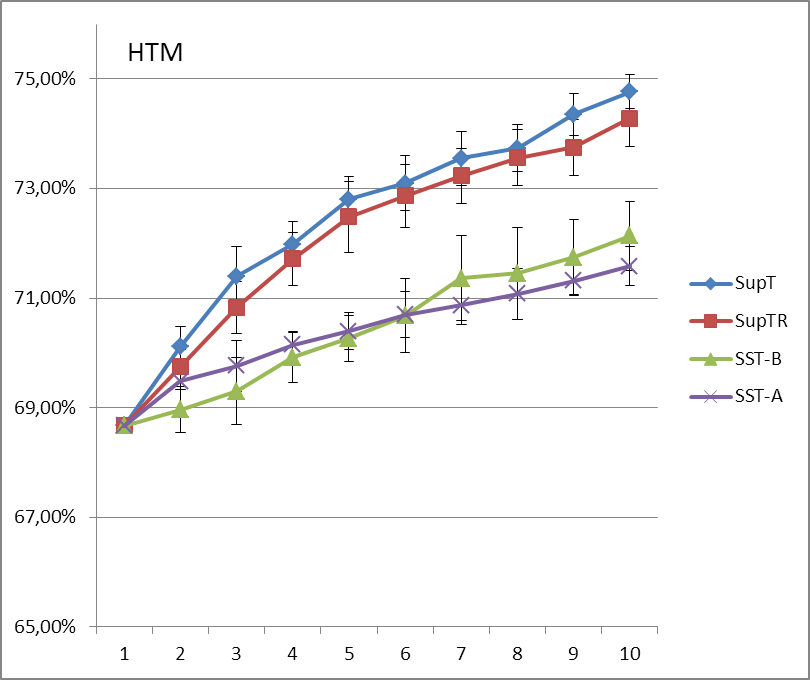}

\end{subfigure}%
\begin{subfigure}{.5\textwidth}
	\includegraphics[width=0.95\linewidth, height=5.56cm]{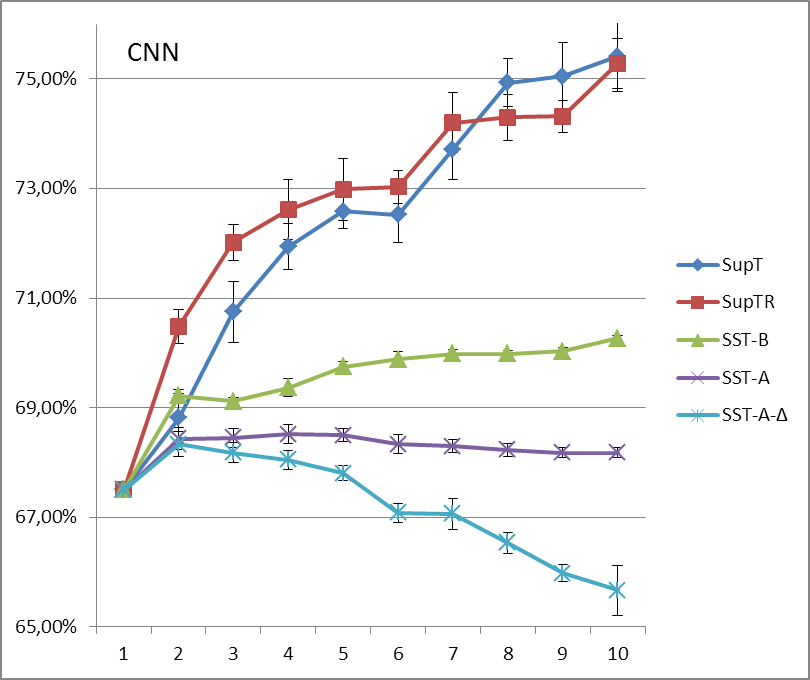}
\end{subfigure}
\caption{HTM and CNN incremental tuning accuracy on \textsf{COIL-100}.}
\label{fig:exp_coil100}
\end{figure}

\textsf{COIL-100} (\cite{nene1996columbia}) contains a larger number of classes than \textsf{NORB} (100 vs 5), but the available variations for each class are much more limited (72 images per class in \textsf{COIL-100} vs 9720 images per class in \textsf{NORB}). The 72 poses of each class are spanned by a single mode of variation (i.e., camera azimuth) which is uniformly sampled with 5 degree steps. The single mode of variation and the limited number of poses make the generation of (disjoint) temporally coherent sequences for incremental learning critical. However, we tried to setup a test-bed close to the \textsf{NORB} one (see Section \ref{sec:norb_revisited}):
\begin{itemize}[leftmargin=.7cm]
\item 6 poses per class (one pose every 60°) are included in the test set; for each test set pose the two adjacent ones (5° degrees before and after) are excluded from the training batches to enforce a \emph{mindist = 2}. 
\item Temporally coherent sequences are obtained for each class by randomly walking the remaining 54 = 72­6 -12 frames. Training batches $TrainB_i$ (1000 patterns wide) are then generated and used for initial supervised training ($TrainB_1$) and successive incremental tuning ($TrainB_2, \dots, TrainB_{10}$). It is worth noting that with respect to the \textsf{NORB} experiments, in this case the forgetting effect induced by incremental tuning is mitigated by an higher overlapping among the tuning batches due to the small number of frames. 
\item To reuse the same HTM and CNN architectures created for \textsf{NORB}, \textsf{COIL-100} images are subsampled (from 128$\times$128) to 32$\times$32 and converted from RGB to grayscale.
\end{itemize}
Fig. \ref{fig:exp_coil100} shows HTM and CNN accuracy for different incremental tuning strategies. We observe that:
\begin{itemize}[leftmargin=.7cm]
\item The trend for supervised strategies is similar to \textsf{NORB}; both HTM and CNN constantly improve initial accuracy as new batches are presented, with CNN slightly overperforming HTM. For HTM regularization seems not providing any advantage, probably due to the shorter sequence length (10 frames here instead of 20 frames in \textsf{NORB}) and the presence of gaps in the sequences (patterns segregated/excluded because of their inclusion in the test set).
\item Here too, semi-supervised strategies performs better for HTM than for CNN. It is worth noting that in this case the base strategy \emph{SST-B} outperforms \emph{SST-A} thus indicating that the self-confidence threshold \emph{sc} (kept fixed at 0.65) is probably too conservative for this dataset. 
\end{itemize}

\section{Discussion and Conclusions}
In this paper we studied semi-supervised tuning based on temporal coherence. The proposed tuning approaches have been evaluated on two deep architectures (HTM and CNN) obtaining partially discordant results. \\
As to HTM our experiments proved that in some conditions even a trivial approach enforcing the output slow change (\textit{SST-B}) can significantly improve classification accuracy. A slightly more complex approach (\textit{SST-A}), exploiting temporal coherence twice: $i)$ to enforce the output slow-change; $ii)$ to compute a self-confidence value to trigger semi-supervised update, proved to be very effective, sometimes approaching the supervised tuning accuracy.\\
Our CNN implementation worked well with supervised tuning strategies, but (unexpectedly) demonstrated a lower capacity to deal with incremental semi-supervised tuning. Of course the encountered limitations could be due to the specific CNN architecture and training, and the outcomes of other recent studies (\cite{goodfellow2013empirical}) can be very useful to check alternative setups (e.g., better investigating the effect of dropout). 
We recognize that the empirical evaluations carried out in this study are still limited, and to validate/generalize our semi-supervised tuning results, we need to test the proposed approaches on other larger datasets, including natural videos of real objects smoothly moving in front of the camera. We plan to collect a new video dataset in the near future and (of course) to work with patterns larger than 32$\times$32 pixels.\\
However, based on the results obtained so far a question emerges: what made HTM more effective than CNN for incremental learning and semi-supervised tuning from temporal coherence? At this stage we do not have an answer to this question, and we can only formulate some hypotheses, by pointing out architectural/training differences that could have a direct impact on forgetting and capability to work with unlabeled data: 
\begin{itemize}[leftmargin=.7cm]
\item \textbf{Pre-training}: \cite{mcrae1993catastrophic} argued that network pre-training can mitigate catastrophic forgetting effects. During initial training HTM self-develop internal memories from patterns of the domain instead of randomly initializing weighs. This could make it more stable and resistant to pattern forgetting and lack of labels. Of course CNN can be pre-trained as well (see \cite{wagner2013learning} for a comparative evaluation of different pre-training approaches), and this is one of directions we intend to follow in our future studies.
\item \textbf{Type of parameters tuned}: CNN training is mostly directed to feature extraction layers (i.e. filter parameters), while HTM + HSR main target are parameters of feature pooling layers. \cite{rehn2014incremental} argued that the most important contribution of HSR is tuning the probabilities denoting how much each coincidence (i.e., a feature extractor) belongs to each group (i.e., a set of feature extractors). Our HTM incremental tuning by HSR is not altering feature extractors, but attempts to optimally arrange existing feature extractors in groups to maximize invariance. Referring to the stability-plasticity dilemma we speculate that keeping feature extractors stable (expecially at low levels) promotes stability while moving pooling parameters is enough to get the required plasticity.
\end{itemize}
In conclusion, we believe that incremental (semi-supervised and unsupervised) tuning, still scarcely studied with deep learning architectures, is a powerful approach to mimic biological learning where continuous (lifelong) learning is a key factor. The lack of supervision, here surrogated by temporal coherence only, can be complemented by other contextual information coming from different modalities (\emph{Multiview learning}), or from different processing paths (e.g., \emph{Co-training}). Of course when supervisor signals are available, both supervised and unsupervised tuning can be fused into an hybrid scheme (as here demonstrated for \textit{SupTR}). The availability of powerful computing platforms, makes the development of continuous learning system feasible for a number of practical applications. For example in our non-optimized HTM implementation, 4 HSR iterations on 1,000 patterns takes about 35 seconds\footnote{On a CPU Xeon W3550 - 4 cores.} we are confident that, upon proper optimization, SST can run on-line once a pre-trained system is switched in working mode.

\bibliography{SST_paper}
\bibliographystyle{iclr2016_conference}

\appendix
\section{Baseline accuracy on NORB}
\label{app:baseline_acc}

Here we report accuracy of HTM and CNN on the ``standard'' normalized-uniform \textsf{NORB} benchmark \cite{lecun2004learning}.\\
We consider monocular 32$\times$32 patterns, and study the classification accuracy on the full test set of 24,300 patterns, for training sets of increasing size. Results reported below are obtained through a 5-fold cross validation, where for each round, 1/5 of the test set was taken as validation set to stop the gradient descent at an optimal point and the remaining 4/5 used to measure accuracy. \\
HTM training was performed as described in \cite{rehn2014incremental}: a subset of the available patterns is used for pre-training and the rest of the patterns for supervised tuning through HSR. This allows to better control the network complexity when scaling to large training sets. Since the HTM pre-training algorithm (\cite{maltoni2011pattern}) internally generates a number of jittered versions\footnote{Consisting of small translations, rotations and scale changes.} of the input patterns to emulate temporally coherent exploration sequences, for a fair comparison we exported these patterns and added them to the training set used for CNN training\footnote{This is not the case for experiment with temporally coherent sequences (reported in section \ref{sec:experiments}.) because when input comes from slowly moving patterns HTM does not need to internally generate jittered patterns.}. The number of HSR iterations (for optimal convergence on the validation set) is almost always less than 50. 
CNN training is performed with mini-batches of 100-200 patterns. The number of error backpropagation iterations (for optimal convergence on the validation set) is almost always less than 150 iterations.

\begin{figure}[h]
\RawFloats
\begin{floatrow}
\ffigbox{%
  \includegraphics[width=\linewidth]{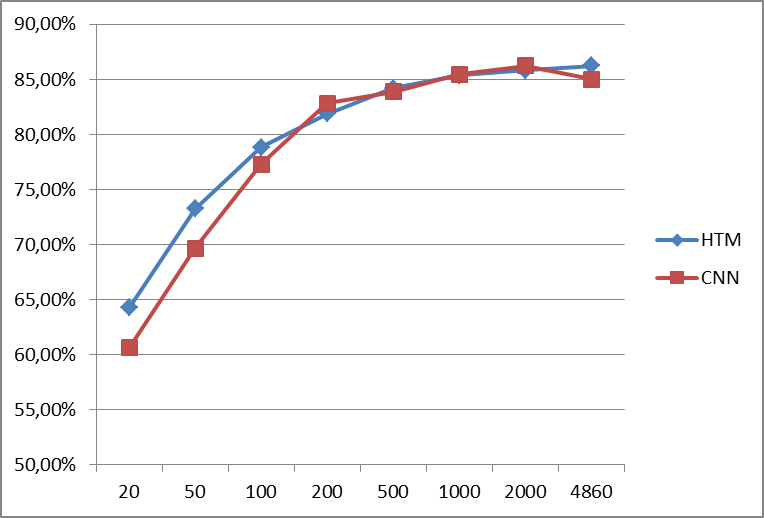}%
}{%
  \caption*{a)}%
}
\capbtabbox{%
\begin{tabular}{llll}
\multicolumn{1}{p{1.4cm}}{\bf Training patterns}  &\multicolumn{1}{p{1.4cm}}{\bf Jittered versions} &\multicolumn{1}{p{1cm}}{\bf HTM} &\multicolumn{1}{p{1cm}}{\bf CNN}
\\ \hline \\
20$\times$5        & 800 & 64.21\% & 60.58\%\\
50$\times$5        & 2,000 & 73.22\% & 69.64\%\\
100$\times$5        & 4,000 & 78.82\% & 77.27\%\\
200$\times$5        & 4,000 & 81.86\% & 82.80\%\\
500$\times$5        & 4,000 & 84.16\% & 83.87\%\\
1,000$\times$5        & 4,000 & 85.37\% & 85.47\%\\
2,000$\times$5        & 4,000 & 85.83\% & 86.20\%\\
4,860$\times$5        & 4,000 & 86.24\% & 85.01\%\\
\end{tabular}
}{%
  \caption*{b)}%
}
\end{floatrow}
\caption{\small HTM and CNN accuracy on standard normalized-uniform \textsf{NORB} benchmark. The labels (number of training patterns per class) in the x-coordinate are equispaced for better readability.}
\label{fig:htm_vs_cnn}
\end{figure}

Fig. \ref{fig:htm_vs_cnn} shows the accuracy of HTM and CNN. When the number of training patterns per class is small (i.e., 20, 50 and 100) HTM accuracy is slightly better than CNN; for larger training sets the accuracy of the two approaches is very similar. Note that with monocular inputs the error is markedly higher with respect to the binocular case reported in section \ref{sec:lenet7_on_theano}.
Concerning the training time, a direct comparison is not possible because of different implementation languages and hardware platforms. In particular, the CNN Theano implementation run on a GPU Tesla C2075 Fermi, while the HTM run on a CPU Xeon W3550 - 4 cores. However, to give a coarse indication, both HTM and CNN training took about 3 hours\footnote{ For a single round of cross-validation.} for the largest training set case: 4860$\times$5 + 4000 patterns. 


\section{Incremental tuning on NORB (native object segregation)}
\label{app:norb_native}

\begin{figure}[h]
\center
\begin{subfigure}{.5\textwidth}
	\includegraphics[width=0.95\linewidth]{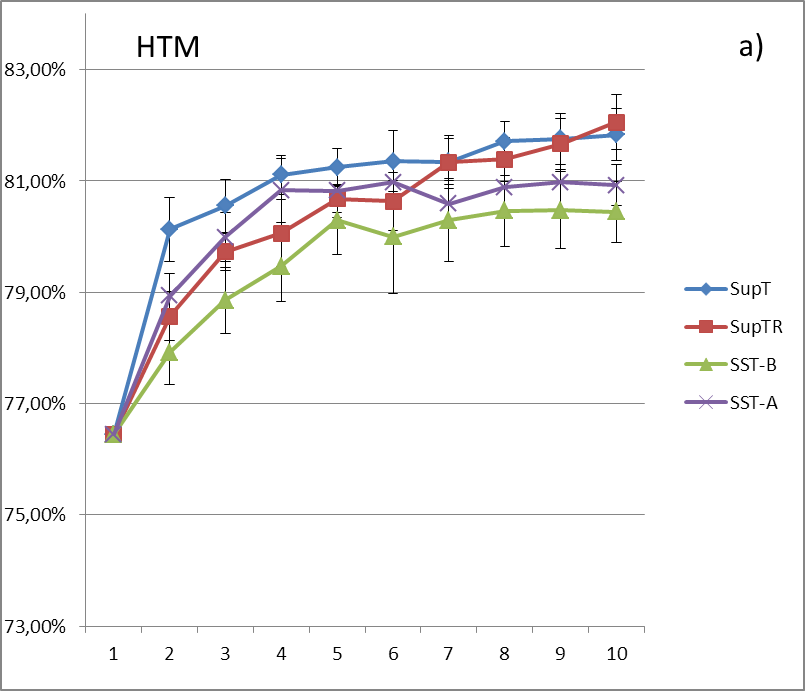}

\end{subfigure}%
\begin{subfigure}{.5\textwidth}
	\includegraphics[width=0.95\linewidth, height=5.56cm]{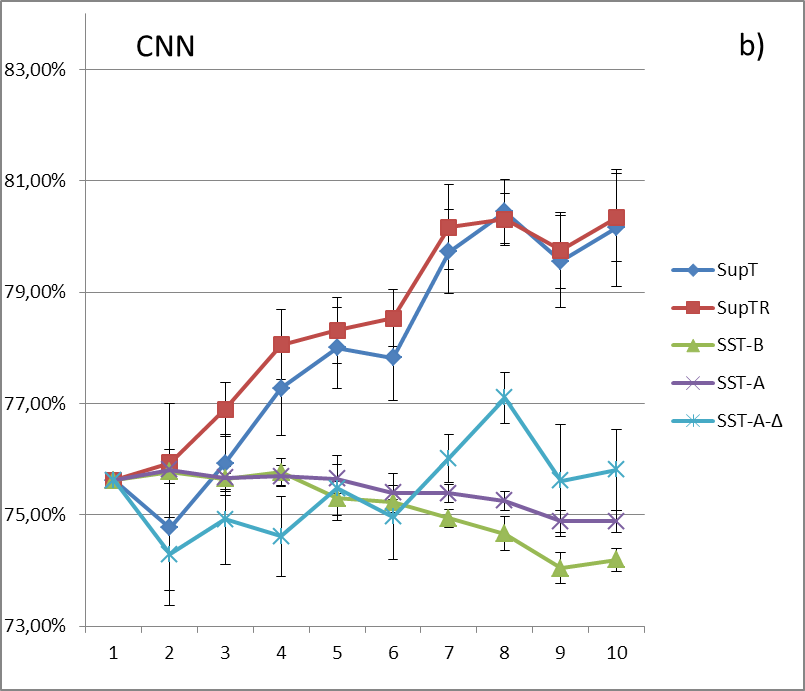}
\end{subfigure}
\caption{HTM and CNN incremental tuning accuracy, when splitting class objects as in the original \textsf{NORB} protocol (for each class: 5 objects in the training set and 5 in the test set). No \emph{mindist} is here necessary between test and training batches because of the object segregation.}
\label{fig:orig_split}
\end{figure}

The \textsf{NORB} benchmark introduced in Section \ref{sec:norb_revisited} focuses on pose and lighting incremental learning and, unlike the original \textsf{NORB} protocol, it does not split the objects in two disjoint groups: for each class, 5 objects in the training set and 5 objects in the test set. Our choice was aimed at isolating the capability of learning pose and invariance from the capability of recognizing different objects of the same class (which is critical in \textsf{NORB} because of the small number of objects per class).
However, to further validate the efficacy of the proposed incremental tuning, here we came back to the native object segregation and report results corresponding to Section \ref{sec:inc_tun} results under this scenario. Figure \ref{fig:orig_split} shows HTM and CNN accuracy for different tuning strategies. We observe that:
\begin{itemize}[leftmargin=.7cm]
\item The trend is very similar to the Section \ref{sec:experiments} experiments: even in this case, supervised strategies work well for both strategies while semi-supervised tuning is effective for HTM but not for our CNN implementation.
\item The accuracy achieved is markedly lower with respect to Section \ref{sec:experiments}, but is in line with results reported in Appendix \ref{app:baseline_acc} if we consider the number of training samples and the forgetting effect due to incremental learning.   
\end{itemize}

\section{HTM overview}
\label{app:htm_overview}

This Appendix provides a brief overview of HTM. A more detailed introduction to HTM structure, forward and backward messaging (including equations) is given in Sections 1 and 2 of \cite{rehn2014incremental}. HTM pre-training algorithms are presented in detail in \cite{maltoni2011pattern} while HTM Supervised Refinement (HSR) is introduced in \cite{rehn2014incremental}.

\paragraph{Structure.}
An HTM has a hierarchical tree structure. The tree is built up by a number of levels, each composed of one or more nodes. A node in one level is bidirectionally connected to one or more nodes in the level below and the number of nodes in each level decreases as we ascend the hierarchy. Conversely, the node receptive fields increase as we move up in the tree structure. By allowing nodes to have multiple parents we can create networks with overlapping receptive fields. The lowest level is the \emph{input} level, and the highest level (with typically only one node) is the \emph{output} level. Levels and nodes in between input and output are called \emph{intermediate} levels and nodes. 
\begin{itemize}[leftmargin=.7cm]
\item Input nodes constitute a sort of interface: in fact, they just forward up the signals coming from the input pattern. 
\item Every intermediate node includes a set, $C$, of so-called coincidence-patterns (or just \emph{coincidences}) and a set, $G$, of coincidence \emph{groups}. A coincidence, $c_i$, is a vector representing a prototypical activation pattern of the node's children. Coincidence groups are clusters of coincidences likely to originate from simple variations of the same input pattern. Coincidences belonging to the same group can be spatially dissimilar but likely to be activated close in time when a pattern smoothly moves through the node's receptive field (i.e., temporal pooling). The assignment of coincidences to groups within each node is encoded in a probability matrix $PCG$, where each element, $PCG_{ji} = P(c_j|g_i)$, represents the probability of a coincidence $c_j$, given a group $g_i$. 
\item The structure of the output node differs from that of the intermediate nodes. In particular the output node has coincidences but not groups. Instead of memorizing groups and group likelihoods, it stores a probability matrix $PCW$, whose elements $PCW_{ji} = P(c_j|w_i)$ represent the probability of coincidence $c_j$ given the class $w_i$.
\end{itemize}

\paragraph{Inference.}
HTM inference (feedforward flow) proceeds from input to output level. Each intermediate node: $i)$ computes its coincidence activations by combining the messages coming from its child-nodes according to the activation patterns encoded by the coincidences themselves; $ii)$ calculates its group activations by mixing coincidence activations through $PCG$ values; finally, $iii)$ passes up information to parent node(s). The output node computes its coincidence activations and turn them to class posterior probabilities according to $PCW$.  

\paragraph{Pre-training.}
HTM pre-training is unsupervised for intermediate levels and partially supervised for the output level. Coincidences are learnt by sampling the space of activation patterns while smoothly moving training patterns across the node(s) receptive fields. Once coincidences are created they are clustered in groups by maximizing a temporal proximity criterion. The output node coincidences are learnt in the same (unsupervised) way but coincidence-class relationships are learnt in a supervised fashion by counting how many times every coincidence is the most active one (i.e., the winner) in the context of each class. 

\paragraph{HTM Supervised Tuning (HSR).}
The probabilities in $PCG$'s (remember there is one $PCG$ matrix for each intermediate node) and $PCW$ are the main elements manipulated by HSR. Similarly to error backpropagation, HSR incrementally updates parameter values by taking steps in direction opposite to the gradient of a loss-function. The whole process is implemented in a simple (and computationally light) way based on native HTM (backward) message passing.

\end{document}